\documentclass[graybox]{svmult}

\usepackage{mathptmx}       
\usepackage{helvet}         
\usepackage{courier}        
\usepackage{type1cm}        
\usepackage{makeidx}         
\usepackage{graphicx}        
\usepackage{multicol}        
\usepackage[bottom]{footmisc}
\usepackage{url}

\usepackage{ifthen}
\usepackage{makeglos}

\graphicspath{{./}{./graphics/}{./graphics/eps/}{./graphics/pdf/}}

\newcommand{\buildbook}{false}

\makeindex             

\makeglossary

\begin{document}

%
%
%

\ifthenelse{\equal{true}{\buildbook}}{
\title{Introduction to Presentation Attack Detection in Iris Biometrics and Recent Advances}
}
{
\title*{Introduction to Presentation Attack Detection in Iris Biometrics and Recent Advances}
}


\author{Aythami Morales, Julian Fierrez, Javier Galbally and Marta Gomez-Barrero}


\institute{Aythami Morales \at School of Engineering, Universidad Autonoma de Madrid, Spain, \email{aythami.morales@uam.es}
\and Julian Fierrez \at School of Engineering, Universidad Autonoma de Madrid, Spain, \email{julian.fierrez@uam.es}
\and Javier Galbally \at European Commission - DG Joint Research Centre, \email{javier.galbally@ec.europa.eu}
\and Marta Gomez-Barrero \at Hochschule Ansbach, Germany, \email{marta.gomez-barrero@hs-ansbach.de}}

%

%
\maketitle


\abstract{
Iris recognition technology has attracted an increasing interest in the last decades in which we have witnessed a migration from research laboratories to real world applications. The deployment of this technology raises questions about the main vulnerabilities and security threats related to these systems. Among these threats presentation attacks stand out as some of the most relevant and studied. Presentation attacks can be defined as presentation of human characteristics or artifacts directly to the capture device of a biometric system trying to interfere its normal operation. In the case of the iris, these attacks include the use of real irises as well as artifacts with different level of sophistication such as photographs or videos. This chapter introduces iris Presentation Attack Detection (PAD) methods that have been developed to reduce the risk posed by presentation attacks. First, we summarise the most popular types of attacks including the main challenges to address. Secondly, we present a taxonomy of Presentation Attack Detection methods as a brief introduction to this very active research area. Finally, we discuss the integration of these methods into Iris Recognition Systems according to the most important scenarios of practical application.
}

\section{Introduction}
\label{sec:1}
The iris\index{Iris} is one of the most popular biometric characteristics within biometric recognition technologies. Since the earliest Daugman publications proposing the iris as a reliable method to identify individuals
~\cite{daugman93irisrecognition} to most recent approaches based on latest machine learning and computer vision techniques~\cite{2019_Access_SurveySuperIris_Alonso,burge13handbookIris,galbally17irischapter}, iris recognition has evolved improving performance, ease of use, and security. Such advances have attracted the interest of researchers and companies boosting the number of products, publications, and applications. The first iris recognition capture devices were developed to work as stand-alone systems~\cite{flom87patent}. However, today iris recognition technology is included as an authentication service in some of the most important operating systems (e.g. Android, Microsoft Windows) and devices (e.g. laptop or desktop computers, smartphones). One-seventh of the world population (1.14 billion people) has been enrolled in the Aadhaar India national biometric ID program~\cite{aadhaar17report} and iris is one of the three biometric characteristics (in addition to fingerprint and face) employed for authentication in that program. The main advantages of iris as a means for personal authentication can be summarised as:

\begin{itemize}
\item{The iris is generated during the prenatal gestation and presents highly random patterns. Such patterns are composed by complex and interrelated shapes and colours. The highly discriminant characteristics of the iris make possible that recognition algorithms reach performances comparable to the most accurate biometric characteristics~\cite{burge13handbookIris}.}
\item{The genetic prevalence on iris is limited and therefore irises from people with shared genes are different. Both irises of a person are considered as different instances, which do not match each other.}
\item{The iris is an internal organ of the eye that is externally visible. It can be acquired at a distance and the advances on capture devices allow to easily integrate iris recognition into portable devices~\cite{2019_HBookSelfie_SuperSelfieFaceIris_Alonso, nguyen17longrangesurvey}.}
\end{itemize}

The fast deployment of iris recognition technologies in real applications has increased the concerns about its security. The applications of iris biometrics include a variety of different scenarios and security levels (e.g. banking, smartphone user authentication, governmental ID programs). Among all threats associated to biometric systems, the resilience against presentation attacks (PAs) emerges as one of the most active research areas in the recent iris biometrics literature. The security of commercial iris systems has been questioned and put to test by users and researchers. For instance, in 2017, the Chaos Computer Club reported their successful attack to the Samsung Galaxy S8 iris scanner using a simple photograph and a contact lens~\cite{chaosclub2017hack}. In the context of biometric systems, Presentation Attacks are defined as presentation of human characteristics or artifacts directly to the capture device of a biometric system trying to interfere its normal operation ~\cite{ISO-spoofing,Galbally2007_Vulnerabilities}. This definition includes spoofing attacks, evasion attacks, and the so called zero-effort attacks. Most of the literature on iris Presentation Attack Detection (PAD) methods is focused on the detection of spoofing attacks. The term liveness detection is also employed in the literature to propose systems capable of classifying between bona fide samples and presentation attack instruments (PAIs) or artifacts used to attack biometric systems. Depending on the motivations of the attacker, we can distinguish two types of attacks:

\begin{itemize}
\item{Impostor: the attacker tries to impersonate the identity of other subject by using his own iris (e.g. zero-effort attacks) or a PAI mimicking the iris of the spoofed identity (e.g. photo, video or synthetic iris). This type of attack requires certain level of knowledge about the iris of the impersonated subject and the characteristics of the iris sensor in order to increase the success of the attack (see Section 2).}
\item{Identity concealer: the attacker tries to evade recognition. Examples in this case include the enrollment of subjects with fake irises (e.g. synthetically generated) or modified irises (e.g. textured contact lens). These examples represent a way to masquerade the real identities.}

\end{itemize}
	
The first PAD approaches proposed in the literature were just theoretical exercises based on potential vulnerabilities~\cite{daugman99irisASpoofing}. In recent years, the number of publications focused on this topic has increased exponentially. Some of the PAD methods discussed in the recent literature have been inspired by methods proposed for other biometric characteristics such as face~\cite{galbally14antispoofingTIP,menotti15irisLDdeep,raghavendra14irisLDcepstral}. However, the iris has various particularities which can be exploited for PAD, such as the dynamic, fast, and involuntary responses of the pupil and the heterogeneous characteristics of the eye’s tissue. The eye reacts according to the amount and nature of the light received. Another large group of PAD methods exploits these dynamic responses and involuntary signals produced by the eye.

This chapter starts by presenting a description of the most important types of attacks from zero-effort attacks to the most sophisticated synthetic eyes. We then introduce iris Presentation Attacks Detection methods and its main challenges. The PAD methods are organised according to the nature of the features employed dividing them into three main groups: hardware-based, software-based, and challenge-response approaches.

The rest of the chapter is organised as follows: Section 2 presents the main vulnerabilities of iris recognition systems with special attention to different types of presentation attacks. Section 3 summarises the Presentation Attacks Detection methods while Section 4 presents the integration of PAD approaches with Iris Recognition Systems. Finally, conclusions are presented in Section 5.

\section{Vulnerabilities in Iris Biometrics}
\label{sec:2}
As already mentioned in the introduction, like any other biometric recognition technology, iris recognition is vulnerable to attacks. Fig.\ref{fig:scheme} includes a typical block diagram of an Iris Recognition System (IRS) and its vulnerable points. The vulnerabilities depend on the characteristics of each module and cover communication protocols, data storage, or resilience against artifact presentations, among others. Several subsystems and not just one will define the security of an IRS:

\begin{figure}[t]
\sidecaption
\includegraphics[scale=.4]{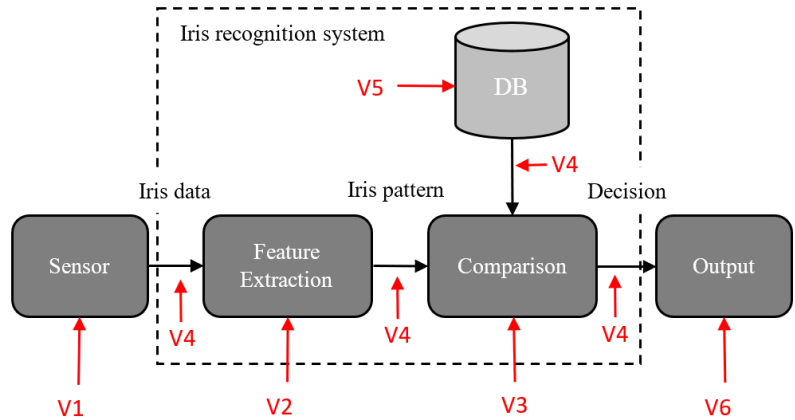}
\caption{Block diagram of a typical Iris Recognition System (IRS) and main vulnerabilities~\cite{ratha01security}.}
\label{fig:scheme}
\end{figure}

\begin{itemize}
\item{Sensor (V1): CCD/CMOS are the most popular sensors including visible and near-infrared images. The iris pattern is usually captured in the form of an image or video. The most important vulnerability is related to the presentation of PAIs (e.g. photos, videos, synthetic eyes) that mimic the characteristics of real irises.}
\item{Feature Extraction and Comparison modules (V2-V3): these software modules comprise the algorithms in charge of pre-processing, segmentation, generation of templates and comparison. Attacks to these modules include the alteration of algorithms to carry out illegitimate operations (e.g. modified templates, altered comparison scores).}
\item{Database (V5): the database is composed by structured data associated to the subject information, devices and iris templates. Any alteration of this information can affect the final response of the system. The security level of the database storage differs depending on the applications. The use of encrypted templates is crucial to ensure the unlinkability between systems and attacks based on weak links.}
\item{Communication channel and actuators (V4 and V6): including internal (e.g. communication between software modules) and external communications (e.g. communication with mechanical actuators or cloud services). The most important vulnerabilities rely on alterations of the information sent and received by the different modules of the IRS.}
\end{itemize}
	
In this work we will focus on presentation attacks to the sensor (V1 vulnerabilities). Key properties of these attacks are their high attack success ratio (if the systems is not properly protected) and the low amount of information about the system needed to perform the attack. Other vulnerabilities not covered by this work include attacks to the database (V5), to the software modules (V2-V3), the communication channels (V4) or actuators at the output (V6). This second group of vulnerabilities require access to the system and countermeasures to these attacks are more related to general system security protocols. These attacks are beyond the scope of this chapter but should not be underestimated.

Regarding the nature of the Presentation Attack Instrument (PAI) employed to spoof the system, the most popular presentation attacks can be divided into the following categories:

\begin{itemize}
\item{Zero-effort attacks: the attack is performed using the iris from the attacker that tries to take advantage of the False Match Rate (FMR) of the system.}
\item{Photo and video attacks: the attack is performed displaying a printed photo, digital image or video from the bona-fide iris directly to the sensor of the IRS.}
\item{Contact lens attacks: the attack is performed using iris patterns printed on  contact lenses.}
\item{Prosthetic  eye attacks: the attack is performed using synthetic eyes generated  to  mimic  the  characteristics  of  real  ones.}
\item{Cadaver eye attacks: the attack is performed using eyes from postmortem subjects.}

\end{itemize}

In the next subsections each of these categories of attacks are discussed in more detail, paying special attention to three important features that define the risk posed by each of the threats: 1) information needed to perform the attack; 2) difficulty to generate the Presentation Attack Instrument (PAI); 3) expected impact of the attack.

\subsection{Zero-effort Attacks}
\label{subsec:21}

In this attack, the impostor does not use any artifact or information about the identity under attack. The iris pattern from the impostor does not match the legitimate pattern and the success of the attack is exclusively related to the False Match Rate (FMR)
 of the system~\cite{hadid15spoofingIntro,johnson12LDframework}. Systems with high FMR will be more vulnerable to this type of attack. Note that the FMR is inversely related to the False Non-Match Rate (FNMR) and both are defined by the operational point of the system. An operation point setup to obtain a low FMR can produce an increment of the FNMR and therefore a higher number of false negatives (legitimate subjects are rejected).
	
\begin{itemize}
\item{Information needed to perform the attack: no information needed.}
\item{Generation of the PAIs: no need to generate a fake iris. The system is attacked using the real iris of the attacker.}
\item{Expected impact of the attack: Most iris recognition systems present very low False Match Rates. The success rate of these attacks can be considered low.}
\end{itemize}

\subsection{Photo and Video Attacks}
\label{subsec:22}
Photo attacks are probably the presentation attacks against IRS most studied in the literature~\cite{czajka13irisSpoofingDB,menotti15irisLDdeep,pacut06irisLD,raghavendra14irisLDlightFieldCamera,virginia08DirectIris,thalheim02directAttacks}. They simply consist on presenting to the sensor an image of the attacked iris, either printed on a sheet of paper or displayed on a digital screen. These attacks are among the most popular ones due to three main factors.

First, with the advent of digital photography and social image sharing (e.g. Flickr, Facebook, Picasaweb and others), headshots of attacked clients from which the iris can be extracted are becoming increasingly easy to obtain. Even though the face or the voice are characteristics more exposed to this new threat, iris patterns can also be obtained from high-resolution face images (e.g. 200 dpi resolution).

\begin{figure}[t]
\sidecaption
\includegraphics[scale=.43]{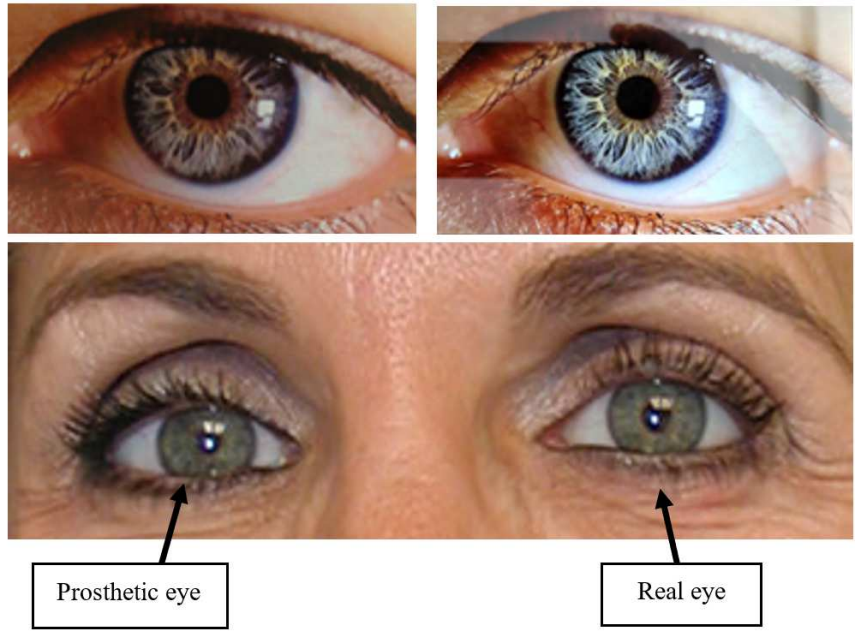}
\caption{Examples of Presentation Attack Instruments (PAIs): printed photo (top-left), screen photo (top-right) and prosthetic eye (bottom) adapted from Soper Brothers and Associates [http://www.soperbrothers.com]}
\label{fig:attacks}
\end{figure}

Second, it is relatively easy to print high quality iris photographs using commercial cameras (up to 12 Megapixels sensors in most of the nowadays smartphones) and ink-printers (1200 dpi in most of the commercial ink-printers). Alternatively, most mobile devices (smartphones and tablets) are equipped with high-resolution screens capable of reproducing very natural images and videos in the visible spectrum (see Fig. \ref{fig:attacks}).

Third, the latest advances in machine learning methods to generate synthetic images open new ways to attack IRS. Recent works demonstrated that it is possible to mimic the pattern of real images using Generative Adversarial Networks \cite{galbally13CVIUreconstructionIris,yadav2019synthesizing}. The synthetic iris can be used to conduct Presentation Attacks or to train PAD methods. 

As a more sophisticated version of photo attacks, the literature has also largely analysed the so-called ``video attacks", that consist on replaying on a digital screen a video of the attacked iris~\cite{he09irisLDwavelets,raja15irisLDvideo,raja15irisLDvideoPhase,zhang11LDirisTexture}. These attacks are able to mimic not ony the static patterns of the iris, but also the dynamic information of the eye that could potentially be exploited to detect static photo attacks (e.g., blinking, dilation/contraction of the pupil). As a limitation compared to the simpler photo attacks, it is more difficult to obtain a video than an image of a given iris.

\begin{itemize}
\item{Information needed to perform the attack: Image or video of the iris of the subject to be impersonated.}
\item{Generation of the PAIs: It is relatively easy to obtain high resolution face photographs from social media and internet profiles. Other options include capturing it using a concealed camera. Once a photo is obtained, if it is of sufficient quality, the iris region can be printed and then presented to the iris acquisition sensor. A screen could also be used for presenting the photograph to the sensor. Another way to obtain the iris would be to steal the raw iris image acquired by an existing iris recognition system in which the subject being spoofed was already enrolled. In the case of video attacks, the availability of high resolution videos of the iris is rather low in comparison with photographs. Nonetheless, it is relatively easy to capture a video in public spaces and the usage of long range sensors open allows video capturing at a distance with high resolution qualities.  }
\item{Expected impact of the attack: The literature offers a large number of approaches with good detection rates of printed photo attacks~\cite{czajka13irisSpoofingDB,menotti15irisLDdeep,pacut06irisLD,raghavendra14irisLDlightFieldCamera,virginia08DirectIris,yambay14livdetIris2013}. However, most of these methods exploit the lack of realism/quality of printed images in comparison with bona fide samples. The superior quality of new screens capable of reproducing digital images and video attacks with high definition represent a difficult challenge for PAD approaches based on visible spectrum imaging. To overcome this potential threat, new commercial scanners based on near-infrared cameras or even on 3D sensors have recently been proposed.}
\end{itemize}

\subsection{Contact Lens Attacks}
\label{subsec:23}
This type of attack uses contact lenses created to mimic the pattern of a different individual (impostor attack) or contact lenses created to masquerade the identity (identity concealer attack). The case of a potential identity concealer attack is particularly worrying because nowadays more and more people wear contact lenses (approximately 125 million people worldwide wear contact lenses). We can differentiate between transparent contact lenses (used in general to correct some sight handicap like myopia) and textured contact lenses (also known as printed). Textured contact lenses change the original iris information by the superposition of synthetic patterns. Although these contact lenses are mostly related to cosmetic applications, the same technology can be potentially used to print iris patterns from real subjects in order to carry out impostor attacks. If an individual is enrolled into the IRS without taking off the textured contact lenses, the IRS can be compromised at a later stage. Note that asking to remove transparent/corrective contact lenses before enrolment or recognition is a non-desirable solution as it clearly decreases the user comfort and usability.
	
\begin{itemize}
\item{Information needed to perform the attack: Image of the iris of the client to be attacked for impostor attacks. No information needed for indentity concealer attacks.}
\item{Generation of the PAIs: In comparison with photo or video attacks, the generation of textured contact lenses requires a more sophisticated method based on optometrist devices and protocols.}
\item{Expected impact of the attack: These type of attacks represent a great challenge for either automatic PAD systems or visual inspection by humans. It has been reported by several researchers that it is actually possible to spoof iris recognition systems with well-made contact lenses~\cite{daugman04irisLD,he09irisLDwavelets,seelen05AntispoofingContactLenses,wei08irisLenses,zhang10LDirisLBP,zhang11LDirisTexture}.}
\end{itemize}

\subsection{Synthetic Eye Attacks}
\label{subsec:24}
This type of attack is the most sophisticated. Prosthetic eyes have been used since the beginning of 20th century to reduce the esthetic impact related to the absence of eyes (e.g. blindness, amputations, etc.). Current technologies for prosthetic manufacturing allow mimicking the most important attributes of the eye with very realistic results. The similarity goes beyond the visual appearance including manufacturing materials with similar physical properties (e.g. elasticity, density). The number of studies including attacks to iris biometric systems using synthetic eyes is still low~\cite{lefohn03IrisOcularist}.
	
\begin{itemize}
\item{Information needed to perform the attack: Image of the eye of the client to be attacked.}
\item{Generation of the PAIs: this is probably the most sophisticated attack method as it involves the generation of both 2D images and 3D structures. Manually made in the past, 3D-printers and their application to the prosthetic field have revolutionised the generation of synthetic body parts.}
\item{Expected impact of the attack: Although the number of studies is low, the detection of prosthetic eyes represents a big challenge. The detection of these attacks by techniques based on image features is difficult. On the other hand, PAD methods based on dynamic features can be useful to detect the unnatural dynamics of synthetic eyes.}
\end{itemize}

\subsection{Cadaver Eye Attacks}
\label{subsec:25}

Post-mortem biometric analysis is very common in forensic sciences. However, during the last years, this field of study attracted the interest of other research communities. As a result of this increasing interest, researchers have evaluated the potential of attacks based on eyes from post-morten subjects~\cite{trokielewicz2018presentation, trokielewicz2020post}. This type of attack has been included in recent iris PAD competitions demonstrating its challenge~\cite{das202iris}.   

\begin{itemize}
\item{Information needed to perform the attack: no information needed.}
\item{Generation of the PAIs: this type of attack is particularly difficult to perform as the attacker needs to have access to the cadaver of the person to be impersonated.}
\item{Expected impact of the attack: Although the number of studies is low, the detection accuracy of this attack increases with respect to the time gap to the post-mortem time horizon. As a living tissue, the pattern on the iris degrades over time once the subject is dead.}
\end{itemize}

Table \ref{tab:literature} summarises the literature on iris presentation attacks including the most popular public databases available for research purposes.

\begin{table}[t]
\centering
\caption{\textbf{Literature on presentation attack detection methods.} Summary of the literature concerning iris Presentation Attack Detection (PAD) methods depending on the Presentation Attack Instrument (PAI).}
\label{tab:literature}       
%
%
\begin{tabular}{|c|c|c|c|}
\hline\noalign{\smallskip}
Ref. & PAI &  PAD &  Database  \\
\noalign{\smallskip}\svhline\noalign{\smallskip}

\cite{galbally14antispoofingTIP} & Photo & Quality measures & Public \\
\cite{he08irisLDfft} & Photo & Wavelet measures & Proprietary \\
\cite{menotti15irisLDdeep} & Photo & Deep features & Public \\
\cite{czajka15irsLDpupilDynamics} & Video & Pupil dynamics & Proprietary \\
\cite{komogortsev13irisLDocculomotor} & Video & Oculomotor Plant Char. & Proprietary \\
\cite{kanematsu07irisLDbrightness} & Video & Pupillary reflex & Proprietary  \\
\cite{czajka2019iris} & Contact Lens & Photometric features & Public\\
\cite{fang2020robust} & Contact Lens & 2D \& 3D features & Public\\
\cite{zhang10LDirisLBP} & Contact lens & LBP features & Proprietary\\
\cite{chen12LDirisMultispectral} & Various & Hyperspectral features & Proprietary\\
\cite{lee08irisLDpurkinje} & Various & Pupillary reflex & Proprietary \\
\cite{LivDet2017} & Various & Various & Public \\

\cite{fang2021iris} & Various & Deep features & Public\\
\cite{yadav2021cit} & Various & Deep features & Public\\
\cite{yadav2019synthesizing} & Various & Deep features & Public\\
\cite{sharma2020d} & Various & Deep features & Public\\

\noalign{\smallskip}\hline\noalign{\smallskip}
\end{tabular}
\end{table}

\section{Presentation Attack Detection Approaches}
\label{sec:3}

These methods are also know in the literature as liveness detection, anti-spoofing, or artefact detection among others. The term Presentation Attack Detection (PAD) was adopted in the ISO/IEC 30107-1:2016~\cite{ISO-spoofing} and it is now largely accepted by the research community.

The different PAD methods can be categorised according to several characteristics. Some authors propose a taxonomy of PAD methods based on the nature of both methods and attacks: passive or active methods employed to detect static or dynamic attacks~\cite{czajka15irsLDpupilDynamics}. Passive methods include those capable of extracting features from samples obtained by traditional iris recognition systems (e.g. image from the iris sensor). Active methods modify the recognition system in order to obtain features for the PAD method (e.g. dynamic illumination, challenge-response). Static attacks refer to those based on individual samples (e.g. image) while dynamic attacks include PAIs capable of changing with time (e.g. video or lens attacks).

\begin{figure}[t]
\sidecaption
\includegraphics[scale=.3]{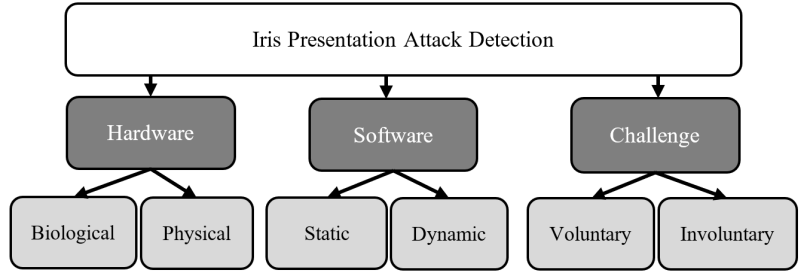}
\caption{Taxonomy of Iris Presentation Attack Detection methods}
\label{fig:taxonomy}
\end{figure}

In this chapter, we introduce the most popular PAD methods according to the nature of the features used to detect the forged iris: hardware-based, software-based, and challenge-response. The challenge-response and most of the hardware methods can be considered active approaches, as they need additional sensors or collaboration from the subject. On the other hand, most of the software methods employ passive approaches in which PAD features are directly obtained from the biometric sample acquired by the iris sensor. Fig. \ref{fig:taxonomy} presents a taxonomny of the iris PAD methods introduced in this chapter.

\subsection{Hardware-based Approaches}
\label{subsec:31}

Also known as sensor-based approaches in the literature. These methods employ specific sensors (in addition to the standard iris sensor) to measure biological and physical characteristics of the eye. These characteristics include physical properties of the eye (e.g. light absorption of the different eye layers or electrical conductivity), and biological properties (e.g. density of the eye tissues,  melanin or blood vessel structures in the eye). The most popular hardware-based PAD approaches that can be found in the literature belong to one of three groups:

\begin{itemize}
\item{Multispectral imaging~\cite{he10LDirisOpticalFeatures,lee06irisLDpurkinje,lee07irisLDreflectanceJournal,park05irisLDreflectanceConf}: The eye includes complex anatomical structures enclosed in three layers. These layers are made of organic tissue with different spectrographic properties. The idea underlying these methods is to use the spectroscopic print of the eye tissues for PAD. Non-living tissue (e.g. paper, crystal from the screens or synthetic materials including contact lenses) will present reflectance characteristics different to those obtained from a real eye. These approaches exploit the use of illumination with different wavelengths that vary according to the method proposed and the physical or biological characteristic being measured (e.g. hemoglobin presents an absorption peak in Near-Infrared Bands).}
\item{3D imaging~\cite{lee10irisLD3d,raghavendra14irisLDlightFieldCamera}: The curvature and 3D nature of the eye has been exploited by researchers to develop PAD methods. The 3D profile of the iris is captured in~\cite{lee10irisLD3d} by using two Near-Infrared light sources and a simple 2D sensor. The idea underlying the method is to detect the shadows on real irises produced by non uniform illumination provided from two different directions. Light Field Cameras (LFC) are used in~\cite{raghavendra14irisLDlightFieldCamera} to acquire multiple depth images and detect the lack of volumetric profiles of photo attacks.}
\item{Electrooculography~\cite{robert12electrooculography}: The electric standing potential between the cornea and retina can be measured and the resulting signal is known as electrooculogram. This potential can be used as a liveness indicator but the acquisition of these signals is invasive and includes the placement of at least two electrodes in the eye region. Advances on non-intrusive new methods to acquire the electrooculogram can boost the interest on these approaches.}
\end{itemize}

\subsection{Software-based Approaches}
\label{subsec:32}

Software-based PAD methods use features directly extracted from the samples obtained by the standard iris sensor. These methods exploit pattern recognition techniques in order to detect fake samples. Techniques can be divided into static or dynamic depending on the nature of the information used. While static approaches search for patterns obtained from a single sample, dynamic approaches exploit time sequences or multiple samples (typically a video sequence).

Some authors propose methods to detect the clues or imperfections introduced by printing devices used during manufacturing of PAIs (e.g. printing process for photo attacks). These imperfections can be detected by Fourier image decomposition~\cite{czajka13irisSpoofingDB,he08irisLDfft,pacut06irisLD}, Wavelet analysis~\cite{he09irisLDwavelets}, or Laplacian transform~\cite{raja15irisLDvideo}. All these methods employ features obtained from the frequency domain in order to detect artificial patterns in fake PAIs. Other authors have explored iris quality measures for PAD. The quality of biometric samples has a direct impact on the performance of biometric systems. The literature includes several approaches to measure the quality of  image-based biometric samples. The application of quality measures as PAD features for iris biometrics has been studied in~\cite{galbally14antispoofingTIP,galbally12LDirisQ}. These techniques exploit iris and image quality in order to detect photo attacks from real irises.

Advances on image processing techniques have also allowed to develop new PAD methods based on the analysis of features obtained at pixel level. These approaches include features obtained from gray level values~\cite{he07irisLDSVM}, edges~\cite{wei08irisLenses}, or color~\cite{alonso14irisLDnirVSvisible}. The idea underlying these methods is that the texture of manufacturing materials show different patterns due to the non-living properties of materials (e.g. density, viscosity). In this line, the method proposed in~\cite{alonso14irisLDnirVSvisible} analyzes image features obtained from near-infrared and visible spectrums. Local descriptors have been also used for iris PAD: Local Binary Patterns~\cite{gragnaniello15irisLDlocalDescriptors,gupta14irisSpoofingPrintDB,he09LDirisBLBP,zhang10LDirisLBP}, Binary Statistical Image Features~\cite{gragnaniello15irisLDlocalDescriptors,raghavendra14irisLDcepstral}, Scale Invariant Feature Transform~\cite{gragnaniello15irisLDlocalDescriptors,sun14irisLDvisualCodebook,zhang11LDirisTexture}, fusion of 2D and 3D features~\cite{fang2020robust}, and Local Phase Quantization~\cite{gragnaniello15irisLDlocalDescriptors}.

Finally, in~\cite{menotti15irisLDdeep}, researchers evaluated for the first time the performance of deep learning techniques for iris photo attack detection with encouraging results. Then the same group of researchers studied how to use those networks to detect more challenging attacks in~\cite{silva15contactLensDetectionDeep}. Since those initial works based in deep learning, over the past few years, many works have continued to work in that direction, proposing novel PAD methods based on on deep architectures including attention learning~\cite{chen2021explainable,fang2021iris}, adversarial learning~\cite{yadav2019synthesizing,yadav2021cit}, and several approaches based on popular convolutional features ~\cite{el2020deep}.

\subsection{Challenge-Response Approaches}
\label{subsec:33}

These methods analyze voluntary and involuntary responses of the human eye. The involuntary responses are part of the processes associated to the neuromotor activities of the eye while the voluntary behavior are responses to specific challenges. Both voluntary and involuntary responses can be driven by external stimuli produced by the PAD system (e.g. changes in the intensity of the light, blink instructions, gaze tracking during dedicated challenges, etc.). The eye reacts to such external stimuli and these reactions can be used as a proof of life to detect attacks based on photos or videos. In addition, there are eye reactions inherent to a living body that can be measured in terms of signals (e.g. permanent oscillation of the eye pupil called hippus, involuntary eye movements called microsaccades, etc.). These reactions can be considered as involuntary responses non-controlled by the subject. The occurrence of these signals is used as a proof of life.

The pupil reactions in presence of uniform light or lighting events was early proposed in~\cite{daugman04irisLD} for PAD applications and more deeply studied in ~\cite{czajka15irsLDpupilDynamics}. As mentioned above, the hippus are permanent oscillations of the pupil that are visible even with uniform illumination. These oscillations range from 0.3-0.7Hz and decline with age. The PAD methods based on hippus have been explored to detect photo attacks and prosthetic eye attacks~\cite{pacut06irisLD,park06irisLDhippus}. However, the difficulties to perform a reliable detection reduce the performance of these methods. Based on similar principles related to eye dynamics, the use of oculomotor plant models to serve as PAD methods was evaluated in~\cite{komogortsev13irisLDocculomotor}.

The reflection of the light in the lens and cornea produces a well-known involuntary effect named Purkinje reflections. This effect is a reflection of the eye to external illumination. At least four Purkinje reflections are usually visible. The reflections change depending on the light source and these changes can be used for liveness detection~\cite{lee06irisLDpurkinje,lee08irisLDpurkinje}. Simple photo and video attacks can be detected by these PAD methods by simply varying the illumination conditions. However, their performance against contact lens or synthetic eye attacks is not clear due to the natural reflections on real pupils (contact lens or photo attacks with pupil holes) or sophisticated fabrication methods (synthetic eyes).

\section{Integration with Iris Recognition Systems}
\label{sec:4}

PAD approaches should be integrated into the iris recognition systems granting a correct and normal workflow. There are two basic integration schemes~\cite{julian18classifiers}:

\begin{itemize}
\item{Parallel integration: the outputs of the IRS and PAD systems are combined before the decision module. The combination method depends on the nature of the output to be combined (e.g. score level or decision level fusion)~\cite{Biggio17multi}}.
\item{Serial integration: the sample is first analyzed by the PAD system. In case of a legitimate subject, the IRS processes the sample. Otherwise, the detection of an attack will avoid unnecessary recognition and the sample will be directly discarded.}
\end{itemize}

The software-based PAD methods are usually included as modules (serial or parallel) in the feature extraction algorithms. A potential problem associated to the inclusion of PAD software is a delay in the recognition time. However, most PAD approaches based on software methods report a low computational complexity that mitigates this concern. The automatic detection of contact lenses plays an important role in software-based approaches. The effects of wearing contact lenses can be critical in case of textured lenses. In~\cite{bowyer14contactLensDetection} the authors reported that textured lenses can cause the FNMR to exceed 90\%. The detection of contact lenses represents a first step in IRS and specific algorithms have been developed and integrated as a preprocessing module~\cite{bowyer14contactLensDetection,menotti15irisLDdeep,yadav14contactLenses}. The final goal of these algorithms is to detect and to filter the images to remove the synthetic pattern.

\begin{figure}[t]
\sidecaption
\includegraphics[scale=.43]{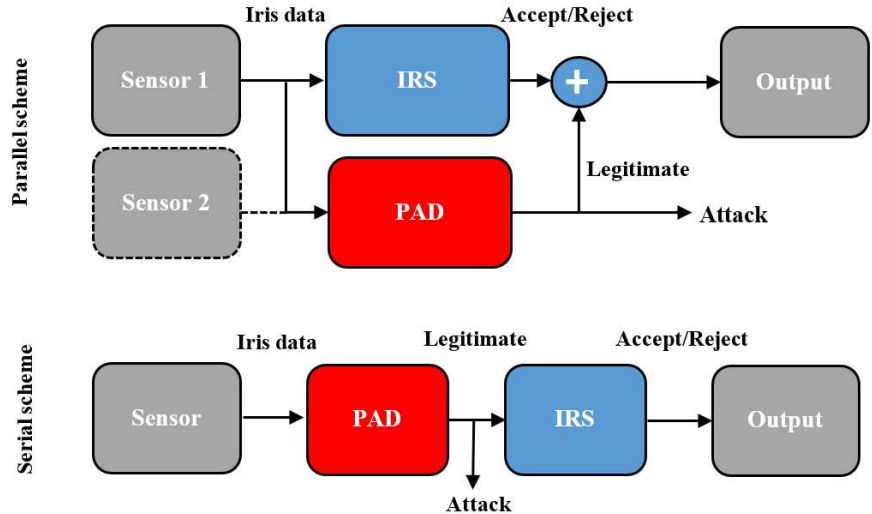}
\caption{Integration of Presentation Attack Detection (PAD) with Iris Recognition Systems (IRS) in parallel (top) and serial (bottom) schemes.}
\label{fig:integration}
\end{figure}

Hardware-based PAD approaches are usually integrated before the iris sensor (serial) or as an independent parallel module (see Fig.~\ref{fig:integration}). In addition to the execution time concerns, hardware-based approaches increase the complexity of the system and the authentication process. Therefore, the main aspects to be analyzed during the integration of those approaches come from the necessity of dedicated sensors and its specific restrictions related to size, time and cost. These are barriers that difficult the integration of hardware-based approaches into mobile devices (e.g. smartphones).

The main drawback of challenge-response approaches is the increased level of collaboration needed from the subject (either for serial or parallel schemes). This collaboration usually introduces delays in the recognition process and some subjects can perceive it as an unfriendly process.

\section{Conclusions}
\label{sec:5}

Iris recognition systems have been improved over the last decade achieving better performance~\cite{das202iris}, more convenient acquisition at a distance~\cite{nguyen17longrangesurvey}, and full integration with mobile devices~\cite{2019_HBookSelfie_SuperSelfieFaceIris_Alonso}. However, the robustness against attacks is still a challenge for the research community and industrial applications~\cite{hadid15spoofingIntro}. Researchers have shown the vulnerability of iris recognition systems and there is a consensus about the necessity of finding new methods to improve the security of iris biometrics. Among the different types of attacks, presentation attacks represent a key concern because of their simplicity and high attack success rates. The acquisition at a distance achieved by recent advances on new sensors and the public exposure of face, and therefore iris, make relatively easy to obtain iris patterns and use them for malicious purposes. The literature on PAD methods is large including a broad variety of methods, databases and protocols. Over the next years, it will be desirable to unify the research community into common benchmarks and protocols. Even if the current technology shows high detection rates for the simplest attacks (e.g. zero-effort and photo attacks), there are still challenges associated to the most sophisticated attacks such as those using textured contact lenses and synthetic eyes.

Future work in this area of iris PAD may exploit: 1) related research being conducted to recognise the periocular region~\cite{alonso2019cross}, 2) other information from the face when the iris image or the biometric system as a whole includes a partial or a full face~\cite{2018_TIFS_SoftWildAnno_Sosa}, and 3) related research in characterizing natural vs artificially-generated faces (DeepFakes) using recent deep learning methods~\cite{2020_INFFUS_SurveyDeepFakes_Tolosana}. All these complementary fields of research can provide valuable information for improved PAD when the iris images or videos are accompanied by additional information from the face surrounding the iris.

\begin{acknowledgement}
This work was mostly done (2nd Edition of the book) in the context of the
TABULA RASA\glossary{TABULA RASA: Trusted Biometrics under Spoofing Attacks.} and
BEAT\glossary{BEAT: Biometrics Evaluation and Testing.}
projects funded under the 7th Framework Programme of EU\index{7th European Framework Programme (FP7)}. The 3rd Edition update has been made in the context of EU H2020 projects PRIMA and TRESPASS-ETN. This work was also partially supported by the Spanish project BIBECA (RTI2018-101248-B-I00 MINECO/FEDER) and by the DFG-ANR RESPECT Project (406880674).
\end{acknowledgement}

\ifthenelse{\equal{false}{\buildbook}}{
\printindex
\printglossary
\bibliographystyle{spmpsci}
\bibliography{references}
}

\end{document}